%% file: egpaper_for_review.tex
\documentclass[journal]{IEEEtran}
\usepackage{cite}
\usepackage{amsmath,amssymb,amsfonts,bm}
\usepackage{graphicx}
\usepackage{textcomp}
\usepackage{bm}
\usepackage{tabularray}
\usepackage[ruled,vlined,linesnumbered]{algorithm2e}
\usepackage{subfigure}
\usepackage{multirow}
\newcommand{\nonl}{\renewcommand{\nl}{\let\nl\oldnl}}
\usepackage{cite}
\usepackage{makecell}

\usepackage{pifont}
\usepackage{adjustbox}

\usepackage{booktabs}
\usepackage{float}
\usepackage{color}

\usepackage[table,xcdraw]{xcolor}
\usepackage{enumerate}

\begin{document}
\title{Integrating Large Pre-trained Models into Multimodal Named Entity Recognition with Evidential Fusion}

\author{Weide Liu, Xiaoyang Zhong, Jingwen Hou, Shaohua Li, Haozhe Huang and Yuming Fang
}


\maketitle

\input{0_Abstract.tex}


\IEEEpeerreviewmaketitle

\input{1_Introduction.tex}

\input{2_Related_Work.tex}
\input{3_Methods.tex}
\input{4_Experiments.tex}
\input{5_Conclusion.tex}

\ifCLASSOPTIONcaptionsoff
  \newpage
\fi

{\small
\bibliographystyle{IEEEtran}
\bibliography{egbib}
}

\end{document}

%% file: 0_Abstract.tex
\begin{abstract}
Multimodal Named Entity Recognition (MNER) is a crucial task for information extraction from social media platforms such as Twitter. Most current methods rely on attention weights to extract information from both text and images but are often unreliable and lack interpretability. To address this problem, we propose incorporating uncertainty estimation into the MNER task, producing trustworthy predictions. Our proposed algorithm models the distribution of each modality as a Normal-inverse Gamma distribution, and fuses them into a unified distribution with an evidential fusion mechanism, enabling hierarchical characterization of uncertainties and promotion of prediction accuracy and trustworthiness. 
Additionally, we explore the potential of pre-trained large foundation models in MNER and propose an efficient fusion approach that leverages their robust feature representations. 
Experiments on two datasets demonstrate that our proposed method outperforms the baselines and achieves new state-of-the-art performance.
\end{abstract}

%% file: 1_Introduction.tex
\section{Introduction}
With the development of social media platforms such as Twitter, multimodal information has become a ubiquitous part of everyday life for many individuals. These platforms serve as important sources for various information extraction applications, including open event extraction and social knowledge graph construction. As a crucial component of these applications, named entity recognition (NER) aims to identify named entities (NEs) and classify them into predefined categories, such as person, location, and organization.
However, Twitter contains multimodal information rather than single-language modal information. To fully utilize the information for each tweet, recent research on tweets has focused on incorporating multimodal learning techniques to enhance linguistic representations through the use of visual cues in tweets. Currently, most of the previous methods~\cite{umt,maf,sun2021rpbert,zhang-UMGF,more} rely on attention weights to extract visual clues related to NEs. However, text and images in tweets may not always be relevant to one another. In fact, some studies~\cite{HVPNet} have shown that text-image relationships where the image does not always match, which may lead to attention-based models producing visual attention even when the text and image are not related, which may negatively impact text inference.

One key limitation of current methods is their lack of reliability and interpretability, rendering them unsuitable for cost-sensitive applications. Furthermore, these models often assume the stability of each modality's quality, leading to inaccurate predictions, especially when features are ambiguous. Additionally, these models tend to make over-confident predictions without accounting for uncertainty. This is a significant problem when each modality produces conflicting decisions due to limited information from a single modality and can be particularly harmful in safety-critical applications.

To address these issues, incorporating uncertainty estimation becomes imperative in multimodal NER prediction to ensure trustworthiness. Models without uncertainty estimation are unreliable as they can be influenced by noise or irrelevant data. Hence, it is crucial to incorporate uncertainty into AI-based systems. Specifically, when a model encounters an image and lacks sufficient confidence to identify the object, it should be able to express ``I am not sure." Models without uncertainty estimation are susceptible to attacks and can lead to costly mistakes in critical decisions.

\input{Figures/Motivation}

As shown in Fig.~\ref{Figure: Motivation}, we propose a novel algorithm for predicting multimodal Named Entity Recognition (MNER) in a trustworthy manner by integrating uncertainty estimation. Our proposed algorithm develops a unified framework that models uncertainty in a fully probabilistic manner. Specifically, we model the distribution of each modality as Normal-Inverse Gamma (NIG) distributions. By fusing the multiple modalities into a new unified NIG distribution, the proposed model utilizes a hierarchically characterized uncertainty to fuse information from different modalities in an evidential fusion way. Ultimately, the fused features promote both prediction accuracy and trustworthiness.

Furthermore, another limitation of current MNER models is the lack of robust feature representation due to limited training samples. However, recent advancements in large foundational models have revolutionized image recognition by exhibiting exceptional performance and generalization capabilities across various images. For instance, the CLIP~\cite{clip} model, trained on an extensive dataset comprising 400 million (image, text) pairs, excels in zero-shot learning and possesses robust representation learning capabilities. Another noteworthy model, the segment anything model~\cite{sam}, is dedicated to segmentation tasks and has been trained on over 1 billion masks, enabling accurate object mask generation based on prompts or in a fully automatic manner. Despite the significant progress made by these large models in vision-language recognition, their potential in the MNER domain remains largely unexplored. In this paper, we aim to explore the applicability of these models in the MNER task and propose an efficient fusion approach that combines their robust feature representations with our proposed evidential fusion method.

In summary, our contributions are summarized as follows:
\begin{itemize}
    \item We propose a novel MNER architecture that enhances the fusion of multiple modalities within an evidential regression framework. The evidential fusion utilizes modality-specific uncertainty to promote trustworthiness.
    \item To the best of our knowledge, we are the first to explore the integration of large foundational models into the MNER task. Specifically, we fuse pre-trained robust feature representations from these models into existing MNER algorithms with our proposed evidential fusion. Through extensive experiments conducted on two datasets and with four baseline models, we validate the effectiveness, robustness, and reliability of our proposed approach in the MNER task.
    \item The performance on Twitter-15 and Twitter-17 datasets demonstrate that our methods outperform the baselines and achieve new state-of-the-art.
\end{itemize}

%% file: Figures/Motivation.tex
 \begin{figure*}[t]
  \centering
    \includegraphics[width=1\linewidth]{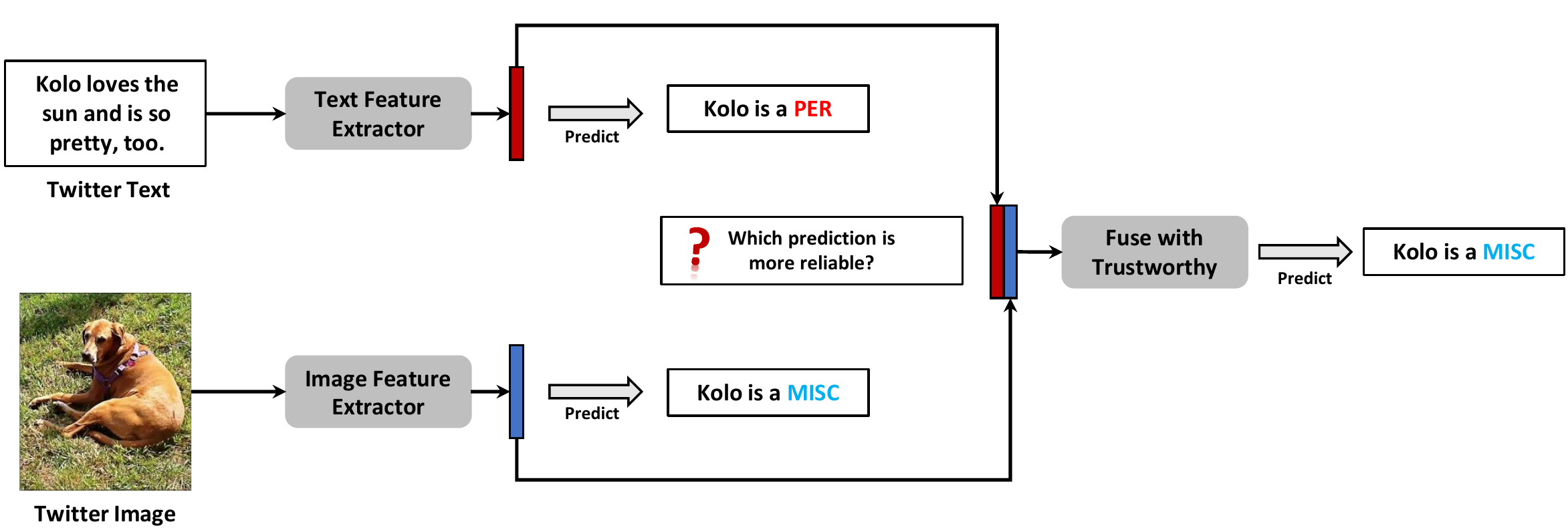}
    \caption{When both an image and its corresponding text are provided as inputs, predicting the final Named Entity Recognition (NER) output can be challenging. Particularly when different modalities produce conflicting decisions due to limited information from a single modality, it becomes difficult to decide which prediction is more reliable and obtain a confidently accurate and stable prediction. To address this issue, we propose to fuse the features from different modalities in an evidential way and provide a trustworthy prediction.}
    \label{Figure: Motivation}
    \vspace{-0.3cm}
\end{figure*}

%% file: 2_Related_Work.tex
\section{Related Work}
\subsection{Named Entity Recognition}
Named Entity Recognition (NER) has been a well-studied problem for decades \cite{Sundheim1995NamedET}. In most research, NER is treated as a sequence labeling problem, and linear-chain CRF is often used to achieve state-of-the-art accuracy \cite{ma2016end,lample2016neural,more}.
Recently, the use of stronger token representations, such as pre-trained contextual embeddings (e.g., BERT \cite{devlin2018bert}, and Flair \cite{akbik2018contextual}, has led to significant improvements in accuracy. The recent method LUKE~\cite{yamada2020luke} has utilized the strengths of pre-trained contextual embeddings in encoding document-level contexts for token representations.
However, in the era of social media, NER alone is no longer sufficient to meet the information extraction needs of users. To address this issue, multimodal NER has been proposed, which utilizes image information to improve the accuracy of related text predictions. In this paper, we aim to study the task of multimodal NER.

\subsection{Multimodal Named Entity Recognition}
Named entity recognition (NER) and relation extraction (RE) are two critical components of information extraction and have received significant attention in the research community~\cite{gcdt,docunet,DBLP:conf/naacl/LiuFTCZHG21,knowprompt,lighter}. While earlier studies mainly focused on textual modality and standard text, the popularity of multimodal data on social media platforms has driven recent research toward developing methods for multimodal named entity recognition (MNER).
In the early stages of Multimodal Named Entity Recognition (MNER) research, several studies proposed to encode the text through RNN and the image through CNN, and then design an implicit interaction to model the information between the two modalities~\cite{zhang2018adaptive,lu2018visual, moon2018multimodal, arshad2019aiding,tip1}.

Recently, with the developemnt of deep learning~\cite{liu2020crnet,liu2020guided,liu2020weakly,liu2021cross,liu2021few,liu2021few_tmm,liu2022crcnet,liu2022long,liu2023harmonizing,zhang2020splitting,hou2022distilling,hou2022interaction,yue2023benchmarking,yue2023perceptual,zhan2023towards}, some studies have proposed leveraging regional image features to represent objects in the image in order to explore fine-grained semantic correspondences between image and text~\cite{yu-etal-2020-improving,zhang-UMGF}. However, most of these methods ignore the error sensitivity of the information extraction task. One exception is the study by ~\cite{sun2021rpbert}, which proposes to learn a text-image relation classifier to enhance the multimodal BERT representation and reduce the interference from irrelevant images. However, this method requires extensive annotation for the irrelevance of image-text pairs. 

Most of these models do not consider the uncertainty in the fusing process of the multimodal information for the final prediction. This uncertainty can result in suboptimal performance or even incorrect predictions. In this paper, we aim to address this issue by introducing a novel approach that integrates uncertainty loss into the MNER task. Our method provides a way to assess and mitigate the uncertainty in the fusing process, leading to improved performance on various multimodal NER datasets.

\subsection{Uncertainty Estimation}
Quantifying uncertainty in machine learning models has been widely studied~\cite{2018Noise, QaddoumReliable}, especially in applications that require high safety and reliability, such as autonomous vehicle control~\cite{Khodayari2010A} and medical diagnosis~\cite{perrin2009multimodal}.

One approach for modeling uncertainty is through Bayesian neural networks~\cite{neal2012bayesian, mackay1992bayesian}, which place a distribution over model parameters and use marginalization to form a predictive distribution. However, due to the ample parameter space of modern neural networks, these models are difficult to infer. Variational Dropout~\cite{2017dropout} has been proposed as a solution by reducing the variance of the gradient estimator, while MC Dropout~\cite{2016drop} is a more scalable alternative that has been applied to downstream tasks~\cite{kendall2017uncertainties, mukhoti2018evaluating}.

Another approach is using deep ensembles~\cite{2018ensemble}, which have shown great performance in both accuracy and uncertainty estimation. However, their high memory and computational costs can be a challenge. To address this, different sub-networks with shared parameters are trained and integrated~\cite{antoran2020depth}.

Deterministic uncertainty methods are designed to directly output uncertainty and address overconfidence~\cite{van2020uncertainty,2020address,Mukhoti2020Calibrating,deep_evidential}. 
Inspired by deep evidential regression~\cite{deep_evidential}, which is designed for single-modal data. In this paper, we aim to develop the uncertainty algorithm with multimodal modal information for the MNER tasks.

\subsection{Large Pre-trained Model}
Contrastive learning is a widely-used pre-training objective for vision models and has also shown promise for vision-language models. Notably, works like CLIP~\cite{clip}, CLOOB~\cite{furst2022cloob}, ALIGN~\cite{align}, and DeCLIP~\cite{declip} leverage contrastive loss to jointly train text and image encoders on large datasets containing {image, caption} pairs. The objective is to map images and texts to the same feature space, minimizing the distance between matching pairs and maximizing it for non-matching ones. 
LiT~\cite{lit} proposes a simple method to fine-tune the text encoder using the CLIP pre-training objective while keeping the image encoder frozen. In contrast, FLAVA~\cite{singh2022flava} combines contrastive learning with other pre-training strategies to align vision and language embeddings.

Another approach in leveraging pre-trained language models for multimodal tasks involves directly integrating visual information into the layers of a language model decoder through cross-attention mechanisms. Models like VisualGPT~\cite{chen2022visualgpt}, and Flamingo~\cite{alayrac2022flamingo} pre-train their models on tasks such as image captioning and visual question-answering. These models aim to balance the capacity for text generation with efficient integration of visual information, which is especially valuable when large multimodal datasets are unavailable. Segmentation anything~\cite{sam}, has applied vision-language models to the segmentation task, which has further explored the potential of vision-language in the new field. However, the potential of large vision-language models in the MNER task remains unexplored. In this paper, we pioneer the fusion of pre-trained features into the MNER task using an evidential fusion approach to improve performance and reliability.

%% file: 3_Methods.tex
\section{Methods}
\subsection{Task Definition}
Given a dataset $\mathcal{D}$ consisting of pairs of text and image, $(x, I)$, where $x = {x_1, \cdots, x_n}$, multimodal Named Entity Recognition (MNER) aims to predict the outputs $y$. Our approach feeds $(x,I)$ into a multimodal recognition module that returns language features, $z_{T}$, and image features $z_{I}$, from the text extract module and an image-based encoder module, respectively.
The textual and visual modules' features are then concatenated for the final prediction. However, the text and the image are not always matched, and it can not be identified which one is more like to be close to the ground truth. To solve this kind of ambiguous situation, our evidential-based module fuses the predictions using a trustworthy technique from the models based on each modality and makes the final prediction, $P(y|z_{T},z_{I})$.

\subsection{Evidential Multimodal Fusion} \label{sec: fusing}
In this section, we will introduce the proposed novel approach to improve the performance of multimodal Named Entity Recognition (MNER) by addressing the issue of lacking trustworthiness and unreliability in current approaches, such as UMT~\cite{umt}, MAF~\cite{maf}, UMGF~\cite{umgf}, and HVPNet~\cite{HVPNet}. These methods only fuse multimodal features directly to calculate the final prediction, without considering the modality-specific uncertainty. This limits their usefulness for trustworthy decision-making, especially when image and text information produce conflicting results.
To address this problem, we propose an approach that captures modality-specific uncertainty and provides a reliable overall uncertainty for the final output. Our approach can distinguish the more reliable modalities for each sample, leading to a more robust integration. Specifically, as shown in Fig.~\ref{Figure: Pipeline}, we propose to fuse multiple modalities at the predictive distribution level. Our algorithm trains deep neural networks to predict the hyper-parameters of higher-order evidential distributions for the modalities and combines them with multiple single mixtures of Normal-inverse Gamma (NIG) distributions. In the following sections, we will provide detailed information about our approach.

\input{Figures/Pipeline}

\subsection{Preliminary Knowledge} \label{sec:preliminary}
The target for a multimodal NER model is to obtain the model parameters $\mathbf{w}$ to maximize the likelihood function of the whole dataset $\mathcal{D}$:
\begin{equation}
    p(\mathbf{y} \mid \mathcal{D}, \mathbf{w}) = p(y_T \mid \mathbf{w}, \mathbf{z}_T) \times p(y_I \mid \mathbf{w}, \mathbf{z}_I). 
    \end{equation}
    
Inspired by the Deep Evidential Learning~\cite{deep_evidential}, the target dataset distribution $ \hat{y} = \{ \hat{y1}, \hat{y2}, ... , \hat{yN} \}$ should follow a normal distribution, but with unknown mean and variance $(\mu, \sigma^2)$.
To model the distribution, $\mu$ and $\sigma^2$ are assumed to be drawn from Gaussian and Inverse-Gamma distributions, respectively:
\begin{align}
 \hat{y} \sim \mathcal{N} (\mu, \sigma^2),\ \mu \sim \mathcal{N} (\delta, \sigma^2\gamma^{-1}),\ \sigma^2 \sim \Gamma^{-1}(\alpha, \beta),
\end{align}
where $\Gamma^{-1}(\cdot)$ is the gamma function, $\delta\in\mathbb{R}$, $\gamma>0$, $\alpha>1$ and $\beta>0$.

With those assumptions, the posterior distribution of the prediction $y$ follows the form of a NIG distribution $\text{NIG}(\delta,\gamma,\alpha,\beta)$. After that, the $\Phi=2\gamma+\alpha$ has been proposed to measure the confidence of the model~\cite{deep_evidential}.

To estimate the uncertainty, following the ~\cite{der2009aleatory}, the predictive uncertainty consists of two parts: epistemic uncertainty (EU) $\mathbb{E}(\sigma^2)$ and aleatoric uncertainty (AU) $\text{Var}(\mu)$:
\begin{align}
\mathbb{E}(\sigma^2) = \frac{\beta}{\alpha-1}, \text{Var}(\mu) = \frac{\beta}{\gamma(\alpha-1)}.
\end{align}

\subsection{Loss}

During the training process, the optimal object~\cite{deep_evidential} should be like:

\begin{equation}
\label{eq:nllloss}
\begin{split}
\mathcal{L}^{NLL}(\mathbf{w})=\frac{1}{2}\log\left(\frac{\pi}{\gamma}\right)- \alpha\log\left(\Omega\right) + \\ \left(\alpha+\frac{1}{2}\right)\log\left((\hat{y} -\delta)^2\gamma+ \Omega\right)+\log \Psi,
\end{split}
\end{equation}
where $\Omega=2\beta(1+\gamma)$ and  $\Psi=\left(\frac{\Gamma(\alpha)}{\Gamma\left(\alpha+\frac{1}{2}\right)}\right)$.

To minimize the incorrect evidence effectiveness, a regularizing term has been added to the likelihood loss:
\begin{equation}
\label{eq:loss_regularize}
\mathcal{L}^{R}(\mathbf{w}) = |\hat{y}-\delta | \cdot (\Phi) = |\hat{y}-\delta | \cdot (2\gamma+\alpha).
\end{equation}

So the loss for the uncertainty estimation should be like:
\begin{equation}
\label{eq:viewloss}
\mathcal{L}(\mathbf{w})_{u}=\mathcal{L}^{NLL}(\mathbf{w})+\lambda\mathcal{L}^R(\mathbf{w}),
\end{equation}
where the coefficient $\lambda$ is a hyper-parameter to balance these two loss terms.

By incorporating the standard multimodal NER classification loss $L(w)_p$ which aims to estimate the accuracy of the predicted NER to the ground truth, we derive the final optimized loss:

\begin{equation}
\label{eq:all_loss}
\mathcal{L}(\mathbf{w})_{all}= \mathcal{L}(\mathbf{w})_{u} + \mathcal{L}(\mathbf{w})_{p}.
\end{equation}

\input{Figures/Quality_Examples}

\subsection{Estimate the uncertainty from the multimodal NIG}
In this section, we will introduce how to fuse the features from multiple modalities at the predictive distribution level for final prediction.
There are two common strategies for multimodal fusion: feature fusion and decision fusion. Most current methods focus on feature fusion which trains a separate model for each modality and combines their features for final prediction. However, those methods are risky when each modality makes a contrary decision due to partial observation from single modality information, which makes an ambiguous decision.

As mentioned above, the prediction from a single modal will follow the NIG distribution, so the challenge for multimodal NER prediction is to integrate these single NIG distributions into a unified NIG. One intuitive way to fuse the multimodal features is the product of experts (PoE) approach. However, the PoE is not suitable as it may result in a distribution that is not a NIG. A simpler alternative is adding NIGs, but this approach does not consider the uncertainties of each NIG and may be impacted by noisy information. To address these limitations, following OLS~\cite{qian2018big}, we use the NIG summation operator to integrate the NIGs and obtain a new NIG distribution. This operator provides an approximate solution as there is no closed-form solution to infer the parameters of the fused NIG distribution.

 \begin{equation}
 \begin{aligned}
 {y} \sim \sum_{m=1}^{M}\frac{1}{M} NIG(\delta_m,\gamma_m,\alpha_m,\beta_m).
  \label{equ:additive}
 \end{aligned}
 \end{equation}

In particular, given two NIG distributions, i.e., $NIG(\delta_t,\gamma_t,\alpha_t,\beta_t)$ for text modality and image modality $NIG(\delta_i,\gamma_i,\alpha_i,\beta_i)$, the summation of the two NIG distributions will be obtained with:
 \begin{equation}
 \begin{aligned}
     NIG(\delta,\gamma,\alpha,\beta) \triangleq NIG(\delta_t,\gamma_t,\alpha_t,\beta_t)\oplus \\
     NIG(\delta_i,\gamma_i,\alpha_i,\beta_i),
 \end{aligned}
 \end{equation}
  where
  \begin{equation}
\centering
  \begin{aligned}
  \label{eq:def1}
\delta={(\gamma_t+\gamma_i)}^{-1}(\gamma_t\delta_t+\gamma_i\delta_i),\quad \quad  \alpha=\alpha_t+\alpha_i+\frac{1}{2},\\
   \gamma=\gamma_t+\gamma_i,\quad \beta=\beta_t+\beta_i+\frac{1}{2}\gamma_1(\delta_t-\delta)^2+\frac{1}{2}\gamma_i(\delta_i-\delta)^2.
   \end{aligned}
  \end{equation}

The NIG summation considers modality confidence through $\gamma_t$ and $\gamma_i$, where a higher value indicates greater confidence in the mean $\delta_m$. The uncertainty estimation, including both aleatoric and epistemic as mentioned in Section~\ref{sec:preliminary}, is reflected by the combination of $\alpha$, $\beta$, and $\gamma$, which are a combination of individual modality uncertainties and the deviation between the final prediction and each modality's prediction. Thus, the final uncertainty is a joint result of modality-specific uncertainty and prediction variations among modalities.

\subsection{Fusing the pre-trained feature into MNER}
As depicted in Fig.~\ref{Figure: Differnet_fusing}, our approach involves extracting image features from two different perspectives when provided with an image. The first perspective is based on a commonly used backbone network (e.g., resnet152), which is optimized using the target dataset to obtain the target-domain knowledge. The second perspective involves extracting image features from a pre-trained large model (e.g., Clip~\cite{clip}, SAM~\cite{sam}, Dinov2~\cite{zhang2022dino}), which parameters are frozen during training. The frozen pre-trained model aims to obtain robust image feature representations, though without domain-specific knowledge. On the other hand, the commonly used backbone network captures domain knowledge but lacks strong feature representation. To fuse these features effectively, we have designed three fusion strategies.

As illustrated in Fig.~\ref{Figure: Differnet_fusing}, we demonstrate three strategies for fusing features: `a', `b', and `c'.
In strategy `a', features from different encoders are directly fused through concatenation.
In strategy `b', the pre-trained image features are treated as an additional aspect of the common image feature. Initially, these features are fused with text features. After passing through the MNER modules, their output features are directly fused using concatenation in the final decoder.
The third approach, `c', involves fusing the output features by comparing their epistemic uncertainty estimations ($\frac{\beta}{\gamma(\alpha-1)}$). We select the features with higher confidence and decode the combined features.
Following each fusion strategy, MNER modules are employed to decode the features and predict uncertainty with confidence.

\input{Tables/UMT}

\input{Figures/Different_Fusion}

\input{Tables/W_Pre-trained}

\input{Tables/Different_Stratage}

\input{Tables/SOTA}

\input{Tables/hype-parameter}

%% file: Figures/Pipeline.tex
 \begin{figure*}[t]
  \centering
    \includegraphics[width=1\linewidth]{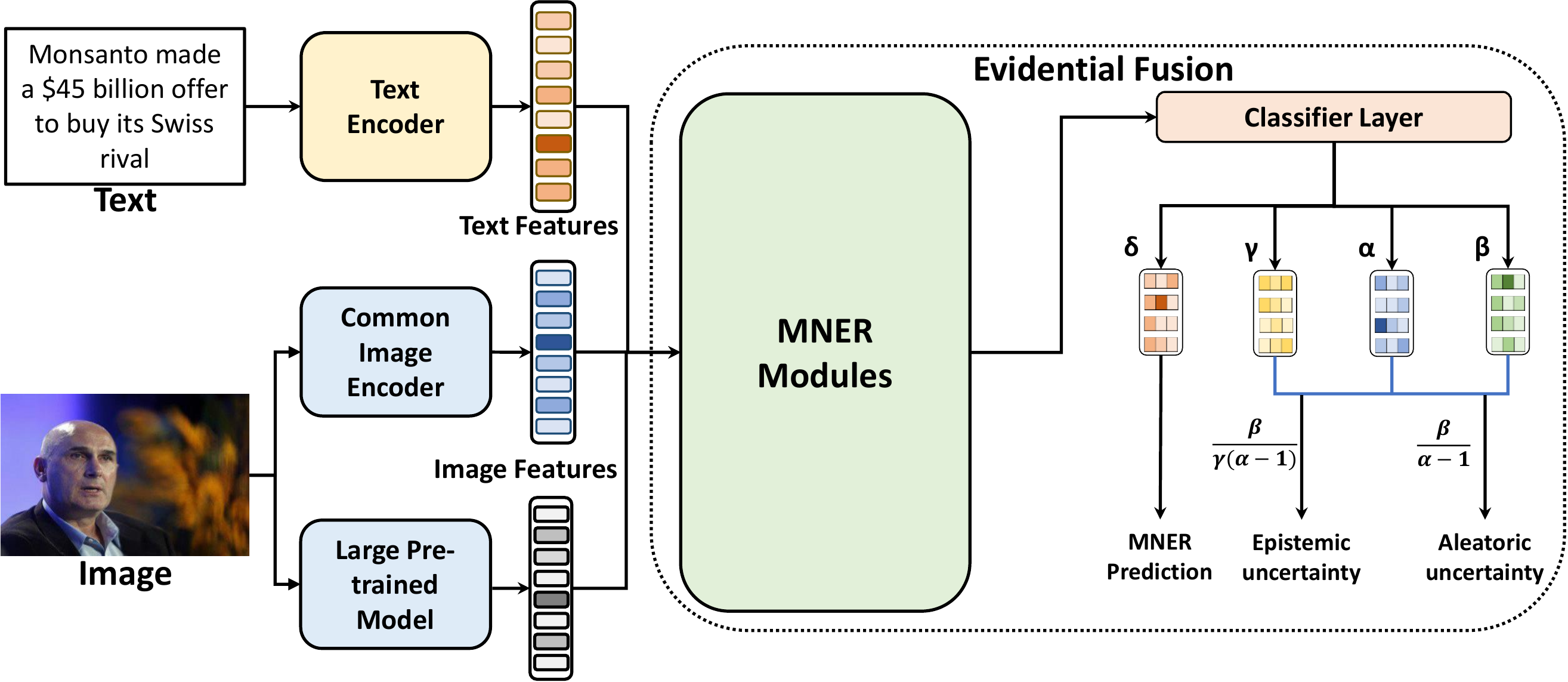}
    \caption{The pipeline of our proposed evidential fusion algorithm. First, we obtain the image and text features from the corresponding encoders and inputs. To fuse the features from different modalities, we use the evidential fusion module, composed of the MNER modules and uncertainty estimation. Instead of directly fusing the features and feeding them into the classifier, our evidential fusion module estimates the prediction and the uncertainty simultaneously. By reducing the model estimate uncertainty, we aim to identify the most confident prediction from different modality features.
    }
    \label{Figure: Pipeline}
\end{figure*}

%% file: Figures/Quality_Examples.tex
 \begin{figure*}[t]
  \centering
    \includegraphics[width=1\linewidth]{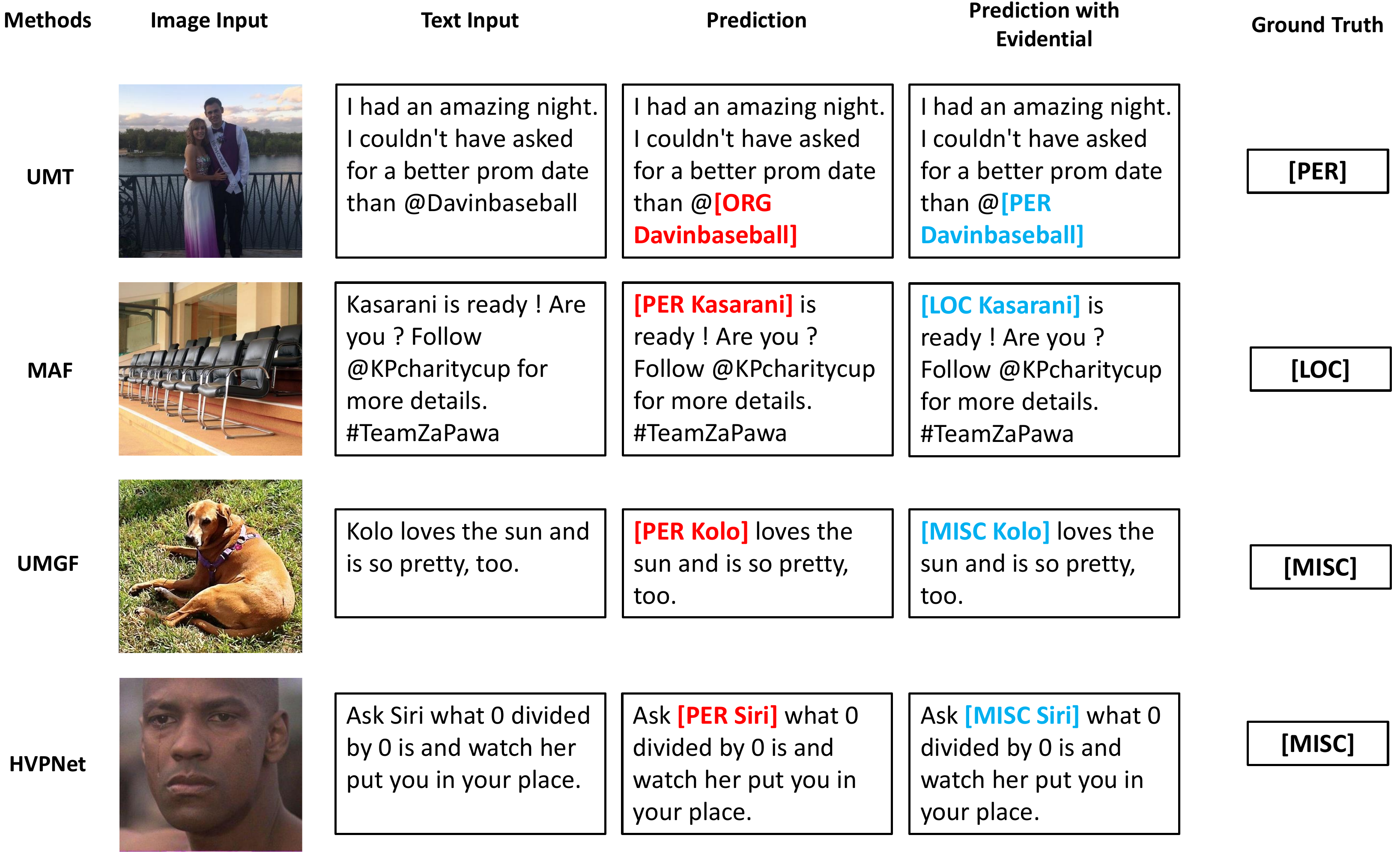}
    \caption{The case studies demonstrate the effectiveness of our evidential fusion approach over existing multimodal NER models, including UMT~\cite{umt}, MAF~\cite{maf}, UMGF~\cite{umgf}, and HVPNet~\cite{HVPNet}. The results clearly indicate the effectiveness of our approach in capturing and utilizing multimodal information, leading to more accurate and trustworthy predictions. Our approach thus provides a more accurate and reliable solution in cases where one modality is ambiguous or two modalities are irrelevant. }
    \label{Figure: Quality_example}
    \vspace{-0.3cm}
\end{figure*}

%% file: Tables/UMT.tex
\begin{table*}[t]

\caption{Ablation study of our proposed evidential fusing with different baselines in two datasets, Twitter-15 and Twitter-17. Our method demonstrates an improvement over the baselines (UMT~\cite{umt}, MAF~\cite{maf}, UMGF~\cite{zhang-UMGF} and HVPNet~\cite{HVPNet}) as evaluated by the metrics of F1, Recall, and Precision. Instead of reporting overall results by averaging across all classes (O(P), O(R), and O(F1)), we also report the performance for each named entity category, including Person (per), Location (loc), Organization (org), and Miscellaneous (misc). The results indicate that our proposed method achieves improvements over most of the categories and baselines on both datasets. $^*$ The denotes the results re-implemented using the official online code.
}

\resizebox{\linewidth}{!}{
\begin{tabular}{l|lllllll|lllllll}
\toprule[1.5pt]
\multicolumn{1}{c|}{\multirow{2}{*}{\textbf{Method}}} & \multicolumn{7}{c}{Twitter-15}                                                & \multicolumn{7}{|c}{Twitter-17}\\  \cline{2-15}
\multicolumn{1}{c|}{}                                 & Per(F1) & Loc(F1) & Org(F1) & Misc(F1) & O(P) & O(R) & O(F1) & Per(F1) & Loc(F1) & Org(F1) & Misc(F1) & O(P) & O(R) & O(F1)\\ \hline

UMT$^*$\cite{umt}                                    
&85.10&81.24&60.97&38.39&72.38&74.38&73.37 
&90.69&85.00&82.67&67.52&83.90&86.01&84.94\\

UMT + Evidential 
&85.46$_{\color{red}{+0.36}}$&81.57$_{\color{red}{+0.33}}$&61.91$_{\color{red}{+0.94}}$&40.80$_{\color{red}{+2.41}}$&71.78$_{\color{blue}{-0.60}}$&75.78$_{\color{red}{+1.30}}$&73.73$_{\color{red}{+0.36}}$
&90.88$_{\color{red}{+0.19}}$&86.78$_{\color{red}{+1.78}}$&83.44$_{\color{red}{+0.77}}$&68.77$_{\color{red}{+1.25}}$&86.70$_{\color{red}{+2.80}}$&84.90$_{\color{blue}{-1.11}}$&85.79$_{\color{red}{+0.85}}$\\  \hline

MAF$^*$\cite{maf}                                         
&84.83&79.84&61.82&40.54&71.61&74.96&73.25 
&91.33&84.44&83.12&66.88&84.70&85.64&85.17\\ 
MAF  + Evidential 
&85.10$_{\color{red}{+0.27}}$&81.47$_{\color{red}{+1.63}}$&61.95$_{\color{red}{+0.13}}$&40.63$_{\color{red}{+0.09}}$&71.93$_{\color{red}{+0.32}}$&75.19$_{\color{red}{+0.23}}$&73.52$_{\color{red}{+0.27}}$
&90.49$_{\color{blue}{-0.84}}$&85.88$_{\color{red}{+1.44}}$&83.33$_{\color{red}{+0.21}}$&69.18$_{\color{red}{+2.30}}$&85.28$_{\color{red}{+0.58}}$&85.34$_{\color{blue}{-0.30}}$&85.31$_{\color{red}{+0.14}}$\\  \hline

UMGF$^*$\cite{umgf}
&86.03&82.66&61.25&40.75&72.69&75.66&74.15 
&92.36&83.84&83.58&66.45&85.48&85.80&85.64 \\  

UMGF + Evidential 
&86.31$_{\color{red}{+0.28}}$&82.38$_{\color{blue}{-0.28}}$&61.60$_{\color{red}{+0.35}}$&41.02$_{\color{red}{+0.27}}$&73.52$_{\color{red}{+1.17}}$&75.61$_{\color{blue}{+0.05}}$&74.55$_{\color{red}{+0.40}}$
&91.81$_{\color{blue}{-0.55}}$&83.31$_{\color{blue}{-0.53}}$&84.13$_{\color{red}{+0.55}}$&69.99$_{\color{red}{+3.54}}$&85.22$_{\color{blue}{-0.26}}$&86.76$_{\color{red}{+0.96}}$&85.99$_{\color{red}{+0.35}}$ \\ \hline

HVPNet-Flat$^*$\cite{HVPNet}
&85.59&82.43&61.94&41.58&73.65&75.52&74.57
&92.39&83.19&84.57&67.09&85.61&86.31&85.96            \\ 
HVPNet-Flat  + Evidential                                  
&86.38$_{\color{red}{+0.79}}$   &81.90$_{\color{blue}{-0.53}}$   &61.31$_{\color{blue}{-0.63}}$   &44.19$_{\color{red}{+2.61}}$   &74.49$_{\color{red}{+0.84}}$   &75.54$_{\color{red}{+0.02}}$   &75.01$_{\color{red}{+0.44}}$   &92.30$_{\color{blue}{-0.09}}$   &84.18$_{\color{red}{+0.99}}$   &84.69$_{\color{red}{+0.12}}$   &67.52$_{\color{red}{+0.43}}$   &85.43$_{\color{blue}{-0.18}}$   &86.82$_{\color{red}{+0.51}}$   &86.12$_{\color{red}{+0.16}}$   
\\  \hline
HVPNet-1T3$^*$\cite{HVPNet}                                           
&86.10&82.44&59.64&43.38&73.82&75.63&74.71
&92.79&84.59&82.71&68.39&85.61&86.31&85.96
\\ 
HVPNet-1T3 + Evidential 
&86.32$_{\color{red}{+0.22}}$   &82.35$_{\color{blue}{-0.09}}$   &61.53$_{\color{red}{+1.89}}$   &44.22$_{\color{red}{+0.84}}$   &74.16$_{\color{red}{+0.34}}$   &76.23$_{\color{red}{+0.60}}$   &75.18$_{\color{red}{+0.47}}$   &92.71$_{\color{blue}{-0.08}}$   &82.72$_{\color{blue}{-1.87}}$   &83.46$_{\color{red}{+0.75}}$   &70.1$_{\color{red}{+1.71}}$   &85.81$_{\color{red}{+0.20}}$   &86.38$_{\color{red}{+0.07}}$   &86.09$_{\color{red}{+0.13}}$   
\\ \hline
HVPNet-OnlyObj$^*$\cite{HVPNet}                                     
&86.58&{83.05}&61.55&42.08&74.85&75.63&75.24
&93.11&83.33&84.51&66.89&85.99&86.75&86.37
\\ 
HVPNet-OnlyObj + Evidential                               
&86.52$_{\color{blue}{-0.06}}$   &82.85$_{\color{blue}{-0.20}}$   &62.51$_{\color{red}{+0.96}}$   &43.09$_{\color{red}{+1.01}}$   &75.4$_{\color{red}{+0.55}}$   &75.67$_{\color{red}{+0.04}}$   &75.54$_{\color{red}{+0.30}}$   &91.95$_{\color{blue}{-1.16}}$   &84.15$_{\color{red}{+0.82}}$   &85.71$_{\color{red}{+1.20}}$   &68.97$_{\color{red}{+2.08}}$   &85.89$_{\color{blue}{-0.10}}$   &86.97$_{\color{red}{+0.22}}$   &86.43$_{\color{red}{+0.06}}$   
\\ \hline
HVPNet$^*$\cite{HVPNet}                                               
&85.96&82.65&63.19&42.17&74.67&75.83&75.24 
&93.15&83.80&85.33&70.16&86.58&87.42&87.00
\\ 
HVPNeT + Evidential                                       
&86.70$_{\color{red}{0.74}}$&82.37$_{\color{blue}{-0.28}}$&63.05$_{\color{blue}{-0.14}}$&45.42$_{\color{red}{+3.25}}$&74.38$_{\color{blue}{-0.29}}$&76.93$_{\color{red}{+1.10}}$&75.63$_{\color{red}{+0.39}}$
&92.97$_{\color{blue}{-0.18}}$&87.01$_{\color{red}{+3.21}}$&86.07$_{\color{red}{+0.74}}$&70.37$_{\color{red}{+0.21}}$&86.69$_{\color{red}{+0.11}}$&88.23$_{\color{red}{+0.81}}$&87.45$_{\color{red}{+0.45}}$\\ \hline

\bottomrule[1.5pt] 
\end{tabular}
}

	\label{tab: UMT}
\end{table*}

%% file: Figures/Different_Fusion.tex
 \begin{figure}[t]
  \centering
    \includegraphics[width=1\linewidth]{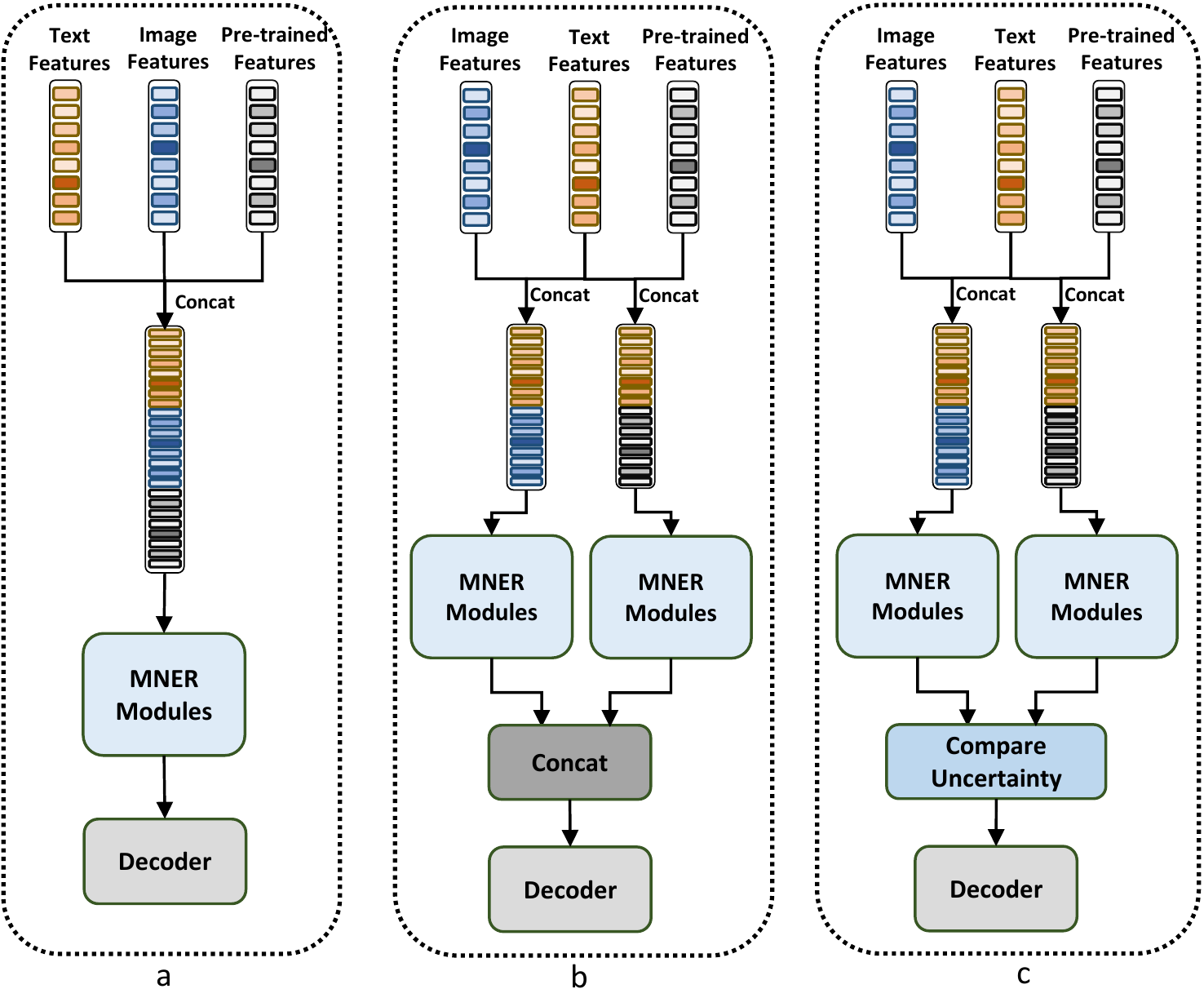}
    \caption{Three ways to fuse the large pre-trained model with image and text features. As depicted in the figures, approach a denotes directly fusing features from different encoders through concatenation. Approach b, on the other hand, treats the pre-trained image features as an additional aspect of the common image feature. These features are first fused with text features, and after passing through the MNER modules, their output features are directly fused using concatenation for the final decoder. The third approach, denoted as c in the figures, involves fusing the output features by comparing their epistemic uncertainty estimations. We select the features with higher confidence and decode the combined features.
    }
    \label{Figure: Differnet_fusing}
\end{figure}

%% file: Tables/W_Pre-trained.tex
\begin{table*}[t]

\caption{Comparison of performance when integrating different types of large pre-trained models, such as the language-vision recognition model CLIP~\cite{clip}, the detection model Dinov2~\cite{zhang2022dino}, and the segmentation model SAM~\cite{sam}. While SAM and Dinov2 do not show any improvement over the baselines, their integration actually leads to a decline in overall performance. On the contrary, the basic CLIP model exhibits a significant improvement over the MNER baselines. Results denoted with $^*$ were re-implemented using publicly available code. 
}

\resizebox{\linewidth}{!}{
\begin{tabular}{l|ccccccc|ccccccc}
\toprule[1.5pt]
\multicolumn{1}{c|}{\multirow{2}{*}{\textbf{Method}}} & \multicolumn{7}{c}{Twitter-15}                                                & \multicolumn{7}{|c}{Twitter-17}\\  \cline{2-15}
\multicolumn{1}{c|}{}                                 & Per(F1) & Loc(F1) & Org(F1) & Misc(F1) & O(P) & O(R) & O(F1) & Per(F1) & Loc(F1) & Org(F1) & Misc(F1) & O(P) & O(R) & O(F1)\\ \hline

UMT$^*$\cite{umt}&                                     
85.10&81.24&60.97&38.39&\textbf{72.38}&74.38&73.37&
90.69&\textbf{85.00}&82.67&\textbf{67.52}&83.90&\textbf{86.01}&84.94\\
UMT + SAM & 
85.04&\textbf{81.35}&61.43&39.32&71.87&74.82&73.32&
\textbf{91.38}&84.66&83.10&64.81&\textbf{86.32}&84.09&85.19 \\  
UMT + Dinov2 & 
84.82&80.87&60.99&38.45&71.02&75.05&72.98&
90.56&84.51&82.93&66.45&84.76&84.83&84.79 \\  
UMT + Clip & 
\textbf{86.82}&79.29&\textbf{62.17}&\textbf{40.17}&71.92&\textbf{75.35}&\textbf{73.59}&
91.26&84.18&\textbf{83.73}&66.89&86.26&84.60&\textbf{85.43} \\   \hline

MAF$^*$\cite{maf}&
84.83&79.84&61.82&40.54&71.61&74.96&73.25&
\textbf{91.33}&84.44&83.12&66.88&84.70&\textbf{85.64}&85.17 \\ 
MAF  + SAM  &
84.67&\textbf{80.86}&61.17&37.15&72.45&74.03&73.23&
90.80&84.27&\textbf{83.25}&66.45&85.80&84.09&84.93 \\  
MAF + Dinov2  &
85.04&80.77&60.53&\textbf{40.71}&71.65&\textbf{75.17}&73.36&
91.05&86.04&82.64&64.88&\textbf{85.94}&84.16&85.04 \\
MAF + Clip  &
\textbf{85.20}&80.66&\textbf{62.18}&38.23&\textbf{72.94}&74.11&\textbf{73.52}&
90.27&\textbf{86.29}&84.12&\textbf{67.10}&85.05&85.49&\textbf{85.27} \\ \hline

UMGF$^*$\cite{umgf}&
\textbf{86.03}&\textbf{82.66}&61.25&\textbf{40.75}&72.69&75.66&74.15&
92.36&83.84&\textbf{83.58}&66.45&85.48&85.80&85.64\\  
UMGF + SAM   &
86.02&82.20&61.68&39.73&72.04&\textbf{76.22}&74.08&
91.85&83.06&83.39&\textbf{66.67}&83.63&\textbf{86.91}&85.24 \\
UMGF + Dinov2 &
85.58&82.42&61.45&39.76&\textbf{72.76}&75.34&74.03&
91.95&84.09&83.35&66.45&84.77&86.02&85.39 \\
UMGF + Clip  &
85.72&81.78&\textbf{63.25}&40.32&72.75&76.08&\textbf{74.38}&
\textbf{92.42}&\textbf{86.44}&82.93&66.38&\textbf{87.40}&84.62&\textbf{85.98} \\ \hline

HVPNet$^*$\cite{HVPNet}&
85.96&82.65&63.19&\textbf{42.17}&74.67&\textbf{75.83}&75.24&
93.15&83.80&85.33&70.16&86.58&87.42&87.00 \\
HVPNet + SAM   &
86.54&82.50&63.24&40.74&74.59&75.77&75.17&
92.62&\textbf{87.31}&84.53&70.10&86.28&87.67&86.97 \\
HVPNet + Dinov2  &
86.29&82.38&61.59&40.50&75.14&74.97&75.06&
91.82&85.96&\textbf{85.86}&70.82&86.63&87.27&86.95 \\
HVPNet + Clip   &
\textbf{86.57}&\textbf{82.76}&\textbf{63.25}&41.07&\textbf{75.29}&75.63&\textbf{75.46}&
\textbf{93.73}&87.25&84.26&\textbf{70.97}&\textbf{86.91}&\textbf{87.93}&\textbf{87.42} \\ \hline

\hline

\bottomrule[1.5pt] 
\end{tabular}
}

	\label{tab: Different_pretrained}
 \vspace{-0.3cm}
\end{table*}

%% file: Tables/Different_Stratage.tex
\begin{table}[t]

\caption{
Comparison of the performance of three feature fusion strategies, denoted as `a', `b', and `c', as illustrated in Fig.~\ref{Figure: Differnet_fusing}. All three strategies resulted in improvements compared to the baseline performance, which included UMT~\cite{umt}, MAF~\cite{maf}, UMGF~\cite{umgf}, and HVPNet~\cite{HVPNet}. Among these strategies, direct fusion of all features (Strategy `a') achieved the best performance in most cases. Results denoted with $^*$ were re-implemented using publicly available code. 
} 

\centering
\resizebox{1\linewidth}{!}{
\begin{tabular}{l|lll|lll}
\toprule[1.5pt]
\multicolumn{1}{c|}{\multirow{2}{*}{\textbf{Methods}}} & \multicolumn{3}{c}{Twitter-15}                                                & \multicolumn{3}{|c}{Twitter-17}\\  \cline{2-7}
\multicolumn{1}{c|}{}
&\multicolumn{1}{c}{O(P)}&\multicolumn{1}{c}{O(R)}&\multicolumn{1}{c|}{O(F1)}
&\multicolumn{1}{c}{O(P)}&\multicolumn{1}{c}{O(R)}&\multicolumn{1}{c}{O(F1)}\\   \hline

UMT$^*$\cite{umt}       &72.38&74.38&73.37&83.90&86.01&84.94\\
UMT + Fusion `a'     &72.41&\textbf{75.37}&\textbf{73.86}&85.77&\textbf{86.27}&\textbf{86.02}\\
UMT + Fusion `b'     &72.12&74.61&73.34&85.58&85.20&85.39\\
UMT + Fusion `c'     &\textbf{72.57}&74.33&73.44&85.30&85.49&85.40\\
\hline
MAF$^*$\cite{maf}       &71.61&74.96&73.25&84.70&85.64&85.17\\
MAF + Fusion `a'    &72.45&\textbf{75.26}&\textbf{73.83}&85.35&\textbf{85.79}&\textbf{85.57}\\
MAF + Fusion `b'    &72.19&74.82&73.48&\textbf{86.45}&84.09&85.25\\ 
MAF + Fusion `c'     &72.27&74.60&73.42&85.33&85.27&85.30\\ 
\hline
UMGF$^*$\cite{umgf      }&72.69&75.66&74.15&85.48&85.80&85.64\\
UMGF + Fusion `a'    &\textbf{74.18}&75.39&\textbf{74.78}&86.18&86.24&\textbf{86.21}\\
UMGF + Fusion `b'    &73.65&75.20&74.42&85.99&85.43&85.71\\
UMGF + Fusion `c'    &73.06&\textbf{76.34}&74.67&85.66&\textbf{86.47}&86.06\\
\hline
HVP$^*$\cite{HVPNet}    &74.67&75.83&75.24&86.58&87.42&87.00\\
HVPNet + Fusion `a'     &\textbf{76.06}&76.00&\textbf{76.01}&\textbf{87.87}&\textbf{87.93}&\textbf{87.90}\\
HVPNet + Fusion `b'     &74.62&76.94&75.76&86.90&87.03&86.96\\
HVPNet + Fusion `c'     &74.52&\textbf{77.28}&75.87&87.17&87.49&87.33\\
\hline
 \bottomrule[1.5pt] 
\end{tabular}
}
\vspace{-0.3cm}
	\label{tab: Different Fusing} 	
\end{table}

%% file: Tables/SOTA.tex
\begin{table}[t]
\centering

\caption{Comparison with previous MNER Methods: our method achieves new state-of-the-art performance. Results denoted with $^*$ were re-implemented using publicly available code. The best performance is highlighted in \textbf{bold}.} 

\resizebox{1\linewidth}{!}{
\begin{tabular}{l|l|l|l|l|l|l|l|l|l|l|l|l|l|l}
\toprule[1.5pt]
\multicolumn{1}{c|}{\multirow{2}{*}{\textbf{Methods}}} & \multicolumn{3}{c}{Twitter-15}                                                & \multicolumn{3}{|c}{Twitter-17}\\  \cline{2-7}
\multicolumn{1}{c|}{}
&\multicolumn{1}{c}{O(P)}&\multicolumn{1}{c}{O(R)}&\multicolumn{1}{c|}{O(F1)}
&\multicolumn{1}{c}{O(P)}&\multicolumn{1}{c}{O(R)}&\multicolumn{1}{c}{O(F1)}\\   \hline

\multicolumn{1}{l|}{AdapCoAtt-BERT-CRF\cite{AdapCoAtt-BERT-CRF}}
&\multicolumn{1}{c}{69.87}&\multicolumn{1}{c}{74.59}&\multicolumn{1}{c|}{72.15}
&\multicolumn{1}{c}{85.13}&\multicolumn{1}{c}{83.20}&\multicolumn{1}{c}{84.10}\\

\multicolumn{1}{l|}{OCSGA\cite{OCSGA}}
&\multicolumn{1}{c}{74.71}&\multicolumn{1}{c}{71.21}&\multicolumn{1}{c|}{72.92}
&\multicolumn{1}{c}{-}&\multicolumn{1}{c}{-}&\multicolumn{1}{c}{-}\\

\multicolumn{1}{l|}{RpBERT\cite{sun2021rpbert}}
&\multicolumn{1}{c}{-}&\multicolumn{1}{c}{-}&\multicolumn{1}{c|}{74.90}
&\multicolumn{1}{c}{-}&\multicolumn{1}{c}{-}&\multicolumn{1}{c}{-}\\   

\multicolumn{1}{l|}{MEGA\cite{MEGA}}
&\multicolumn{1}{c}{70.35}&\multicolumn{1}{c}{74.58}&\multicolumn{1}{c|}{72.35}
&\multicolumn{1}{c}{84.03}&\multicolumn{1}{c}{84.75}&\multicolumn{1}{c}{84.39}\\   

\multicolumn{1}{l|}{VisualBERT\cite{VisualBERT}}
&\multicolumn{1}{c}{68.84}&\multicolumn{1}{c}{71.39}&\multicolumn{1}{c|}{70.09}
&\multicolumn{1}{c}{84.06}&\multicolumn{1}{c}{85.39}&\multicolumn{1}{c}{84.72}\\

\multicolumn{1}{l|}{UMT$^*$\cite{umt}} 
&\multicolumn{1}{c}{72.38}&\multicolumn{1}{c}{74.38}&\multicolumn{1}{c|}{73.37}
&\multicolumn{1}{c}{84.94}&\multicolumn{1}{c}{86.01}&\multicolumn{1}{c}{84.94}\\   

\multicolumn{1}{l|}{MAF$^*$\cite{maf}}
&\multicolumn{1}{c}{71.61}&\multicolumn{1}{c}{74.96}&\multicolumn{1}{c|}{73.25}
&\multicolumn{1}{c}{84.70}&\multicolumn{1}{c}{85.64}&\multicolumn{1}{c}{85.17}\\   
     
\multicolumn{1}{l|}{UMGF$^*$\cite{umgf}}
&\multicolumn{1}{c}{72.69}&\multicolumn{1}{c}{75.66}&\multicolumn{1}{c|}{74.15}
&\multicolumn{1}{c}{85.48}&\multicolumn{1}{c}{85.80}&\multicolumn{1}{c}{85.64}\\  

\multicolumn{1}{l|}{HVPNet$^*$\cite{HVPNet}}
&\multicolumn{1}{c}{74.67}&\multicolumn{1}{c}{75.83}&\multicolumn{1}{c|}{75.24}
&\multicolumn{1}{c}{86.58}&\multicolumn{1}{c}{87.42}&\multicolumn{1}{c}{87.00}\\    \hline

\multicolumn{1}{l|}{Ours}
&\multicolumn{1}{c}{\textbf{76.03}}&\multicolumn{1}{c}{\textbf{76.00}}&\multicolumn{1}{c|}{\textbf{76.01}}
&\multicolumn{1}{c}{\textbf{87.87}}&\multicolumn{1}{c}{\textbf{87.93}}&\multicolumn{1}{c}{\textbf{87.90}}\\

 \bottomrule[1.5pt] 
\end{tabular}
}

	\label{tab: SOTA} 	
\end{table}

%% file: Tables/hype-parameter.tex
\begin{table*}[]
\centering

\caption{Sensitivity analysis for the trade-off parameter $\lambda$. The results show that our method is robust to the hype-parameter setting. Overall,  we can always find a suitable $\lambda$ which brings an obvious improvement over the baseline (The UMT~\cite{umt}, the MAF~\cite{maf}, the UMGF~\cite{umgf} and the HVPNet~\cite{HVPNet}).}

\resizebox{0.7\linewidth}{!}{
\begin{tabular}{c|c|c|c|c|c|c|c}
\toprule[1.5pt]
\multirow{2}{*}{Methods} & \multirow{2}{*}{$\lambda$} & \multicolumn{3}{c|}{Twitter-15} & \multicolumn{3}{c}{Twitter-17} \\ 
 &  & \multicolumn{1}{l}{Overall(P)} & \multicolumn{1}{l}{Overall(R)} & \multicolumn{1}{l|}{Overall(F1)} & \multicolumn{1}{l}{Overall(P)} & \multicolumn{1}{l}{Overall(R)} & \multicolumn{1}{l}{Overall(F1)} \\ \hline
\multirow{5}{*}{UMT} 
& baseline  &\textbf{72.38}&74.38&73.37&83.90&86.01&84.94 \\
&  0.01     &71.78&\textbf{75.78}&\textbf{73.73}&\textbf{86.70}&84.90&\textbf{85.79} \\
&  0.1      &72.26&74.71&73.47&85.35&85.78&85.56 \\
&  1        &71.07&75.57&73.25&84.82&\textbf{86.25}&85.53 \\
&  10       &71.84&73.57&72.69&85.75&84.60&85.17 \\ \hline
\multirow{5}{*}{MAF} 
& baseline  &71.61&74.96&73.25&84.70&\textbf{85.64}&85.17 \\
&  0.01     &\textbf{71.93}&\textbf{75.19}&\textbf{73.52}&85.28&85.34&\textbf{85.31} \\
&  0.1      &71.84&74.71&73.24&\textbf{85.96}&84.44&85.19 \\
&  1        &70.88&74.90&72.84&85.12&85.12&85.12 \\
&  10       &70.99&74.76&72.83&84.73&85.35&85.04 \\ \hline
\multirow{5}{*}{UMGF}
& baseline  &72.69&\textbf{75.66}&74.15&\textbf{85.48}&85.80&85.64 \\
&  0.01     &73.56&74.73&74.14&85.37&86.17&85.77 \\
&  0.1      &73.52&75.61&\textbf{74.55}&85.22&\textbf{86.76}&\textbf{85.99} \\
&  1        &\textbf{73.57}&74.87&74.21&85.32&86.25&85.78 \\
&  10       &72.76&75.09&73.90&85.03&85.47&85.25 \\ \hline
\multirow{5}{*}{HVPNet}
& baseline  &74.67&75.83&75.24&86.58&87.42&87.00 \\
&  0.01     &75.07&75.90&75.48&86.29&87.23&86.76 \\
&  0.1      &\textbf{75.54}&75.71&75.62&\textbf{88.33}&86.20&87.25 \\
&  1        &74.38&\textbf{76.93}&\textbf{75.63}&86.69&88.23&\textbf{87.45} \\
&  10       &75.45&75.54&75.50&85.53&\textbf{88.86}&87.16 \\ 

 \bottomrule[1.5pt]
\end{tabular}
}

\label{tab:hype}
\vspace{-0.2cm}
\end{table*}

%% file: 4_Experiments.tex
\section{Experiment} \label{sec:exp}

\subsection{Dataset and Evaluation Metric}
To validate the efficacy of our proposed method, we conduct evaluations on several multimodal named entity recognition (NER) datasets, including Twitter-15 and Twitter-17. The Twitter-15 dataset includes 4,000 training samples, 1,000 development samples, and 3,357 test samples. The Twitter-17 dataset consists of 3,373 training samples, 723 development samples, and 723 test samples. 

Following previous methods~\cite{more,HVPNet,sun2021rpbert,maf,umt}, we utilize Precision, Recall, and F1 as our evaluation metrics to validate the performance of our method.

\subsection{Effectiveness of Evidential Multimodal Fusion for multimodal NER}

In this section, we integrate our proposed evidential multimodal fusion into four baselines: UMT~\cite{umt}, MAF~\cite{maf}, UMGF~\cite{umgf}, and HVPNet~\cite{HVPNet}. The Unified Multimodal Transformer (UMT) model incorporates an auxiliary text-based entity span detection module and a unified multimodal transformer to enhance entity predictions. MAF incorporates a cross-modal matching module to balance the proportion of visual information, and a cross-modal alignment module for consistency between text and image representations. UMGF proposes a unified multi-modal graph fusion approach to conduct graph encoding based on the graph that can represent the various semantic relationships between multi-modal semantic units. HVPNet proposes the Hierarchical Visual Prefix fusion Network for multimodal NER, enhancing text representations by incorporating visual information as a prefix.

To enhance baseline model performance and improve the trustworthiness of predicted results, we integrate our proposed evidential fusion strategy. In particular, we use four classifiers to predict the uncertainty ($\delta$), conflicting information ($\gamma$), modality-specific uncertainty ($\alpha$), and background information ($\beta$) of the predicted result. These classifiers enable our model to identify the evidential status of predictions and enhance the trustworthiness of the results.

For training our proposed model, we calculate the evidential loss based on the backbone network and the classifier, as described in Equation~\ref{eq:all_loss}. By incorporating the uncertainty loss into the baseline model's original loss function, our proposed model can be trained to improve both the accuracy and trustworthiness of the predicted results. This evidential fusion strategy presents a promising approach to addressing the challenge of multimodal fusion in AI-based systems.

In Table~\ref{tab: UMT}, we demonstrate that our approach outperforms all baselines in terms of F1 scores across all categories for both the Twitter-15 and Twitter-17 datasets. Notably, our evidential fusion yields a 2.41 F1 score improvement over the baseline for the `Misc' category in UMT on the Twitter-15 dataset. Additionally, in the Twitter-17 results, our evidential fusion achieves a 1.78 F1 score improvement for the `Loc' category. 

In the case of MAF, our evidential fusion strategy notably achieves a 1.63 F1 score improvement for the `Loc' category on the Twitter-15 dataset and a 2.30 F1 score improvement for the `Misc' category on the Twitter-17 dataset.

For UMGF, our evidential fusion achieves a 3.54 F1 score improvement for the `Misc' category on the Twitter-17 dataset.

Regarding HVPNet, it is worth noticing that our evidential fusion achieves a 3.25 F1 score improvement for the `Misc' category on the Twitter-15 dataset and a 3.21 F1 score improvement for the `Loc' category on the Twitter-17 dataset.
To further validate the effectiveness of our evidential fusion on HVPNet, we conduct experiments on three variant networks: HVPNet-Flat, HVPNet-1T3, and HVPNet-OnlyObj. Our approach performs well across different objects, demonstrating improvements in various categories and overall metrics.

To better understand how our evidential fusion enhances MNER models, we present several cases in Fig.~\ref{Figure: Quality_example}. Our proposed method significantly improves the accuracy of predictions.

\subsection{Which large pre-trained model will benefit the MNER?}
One critical question that arises is the selection of an appropriate large pre-trained model that can effectively benefit the Multimodal Named Entity Recognition (MNER) task. To explore this, we have integrated various types of large pre-trained models, including the language-vision recognition model CLIP~\cite{clip}, the detection model Dinov2~\cite{zhang2022dino}, and the segmentation model SAM~\cite{sam}. The results of this integration are presented in Table~\ref{tab: Different_pretrained}.

We have integrated these models with the baseline methods such as UMT, MAF, UMGF, and HVPNet. Surprisingly, while SAM and Dinov2 fail to demonstrate any improvement over the baselines. In fact, their integration leads to a decline in the overall performance of the baselines. On the other hand, the basic CLIP model showcases a significant improvement over the MNER baselines.

We hypothesize that the reason behind this contrasting outcome lies in the nature of the MNER task itself. As a multimodal recognition task, MNER primarily focuses on identifying the object within an image rather than determining its specific location. Consequently, the CLIP model, with its emphasis on object recognition, proves to be highly beneficial for the MNER task, resulting in notable improvements. However, the Dinov2 and SAM models, which may excel in other vision-related tasks, do not offer the same advantages for MNER.

\subsection{Fusing the large pre-trained model into MNER}
In this section, we conducted experiments to validate the effectiveness of integrating pre-trained large models using our proposed approach, which employs evidential fusion to enhance the performance of Multimodal Named Entity Recognition (MNER). However, a critical question arises regarding how to fuse the extracted features from these large pre-trained models. Fig.~\ref{Figure: Differnet_fusing} illustrates three strategies we employed for this purpose: direct fusion and separate fusion. The results presented in Table~\ref{tab: Different Fusing} show that all three strategies led to improvements compared to the baseline performance (including UMT~\cite{umt}, MAF~\cite{maf}, UMGF~\cite{umgf} and HVPNet~\cite{HVPNet}). Among them, directly fusing all features (Strategy `a') yielded the best performance in most cases.

\subsection{Comparisons with state-of-the-art methods}
To demonstrate the effectiveness of our proposed approach in the task of multimodal Named Entity Recognition (MNER), we performed a comprehensive comparison with several state-of-the-art methods. 
As shown in Table~\ref{tab: SOTA}, our proposed approach has demonstrated significant improvements in the MNER task, outperforming state-of-the-art methods by incorporating a probabilistic framework with an evidential fusion strategy. 
By modeling the distribution of each modality and the overall output, we were able to produce more accurate and trustworthy predictions. 

\subsection{The uncertainty estimation}
\input{Figures/Uncertainty.tex}

To validate the effectiveness of our proposed evidential fusion on the Twitter-15 and Twitter-17 datasets, we conducted experiments by gradually increasing the training epochs. The aim of this experiment was to determine the ability of the proposed method to estimate epistemic uncertainty. Epistemic uncertainty can be considered as a type of uncertainty that arises from a lack of knowledge or understanding of a system, which can be called ``model uncertainty". In particular, the epistemic uncertainty will be reduced through the model training to understand the system by the model. So the lower of epistemic uncertainty, the more stable and confident the model. In contrast, the aleatoric uncertainty can be considered the variability in a system or the uncertainty of the noise, which cannot be reduced even with additional information or data, or model training. So we do not utilize the aleatoric uncertainty as the metric to validate the stability of the model in this paper.

The results in Fig.~\ref{Figure: uncertainty} demonstrate that the overall epistemic uncertainty steadily declines as the number of training epochs increases. This finding is consistent with the intuitive view that our evidential fusion method helps to decrease uncertainty and produce trustworthy predictions.

During the training process, there may be fluctuations at the beginning, which cause the entire uncertainty to rise initially. However, after the network stabilizes, the uncertainty decreases, indicating that evidential fusion is useful and produces more trustworthy results. Notably, the experiments conducted on the MAF, UMGF and HVPNet with evidential fusion show a sharp and steady decrease in uncertainty on both datasets. In contrast, for the UMT models on the Twitter-17 dataset, there were some fluctuations at first, but the uncertainty eventually stabilized.

By incorporating uncertainty estimation into the MNER task, we promote both prediction accuracy and trustworthiness, making the proposed method suitable for cost-sensitive applications where the reliability and interpretability of predictions are critical.

\subsection{The effectiveness of the hyper-parameter}
The choice of hype-parameter $\lambda$ in equation~\ref{eq:viewloss} is critical for the proposed method's performance. 
As shown in Table~\ref{tab:hype}, we set the hype-parameter $\lambda$ to values ranging from 0.01 to 10. 
Experiments are conducted to investigate the impact of this hyper-parameter on the overall performance, and the results show that the performance is relatively stable across a range of hyper-parameter values. 

%% file: Figures/Uncertainty.tex
 \begin{figure}[t]
  \centering
    \includegraphics[width=1\linewidth]{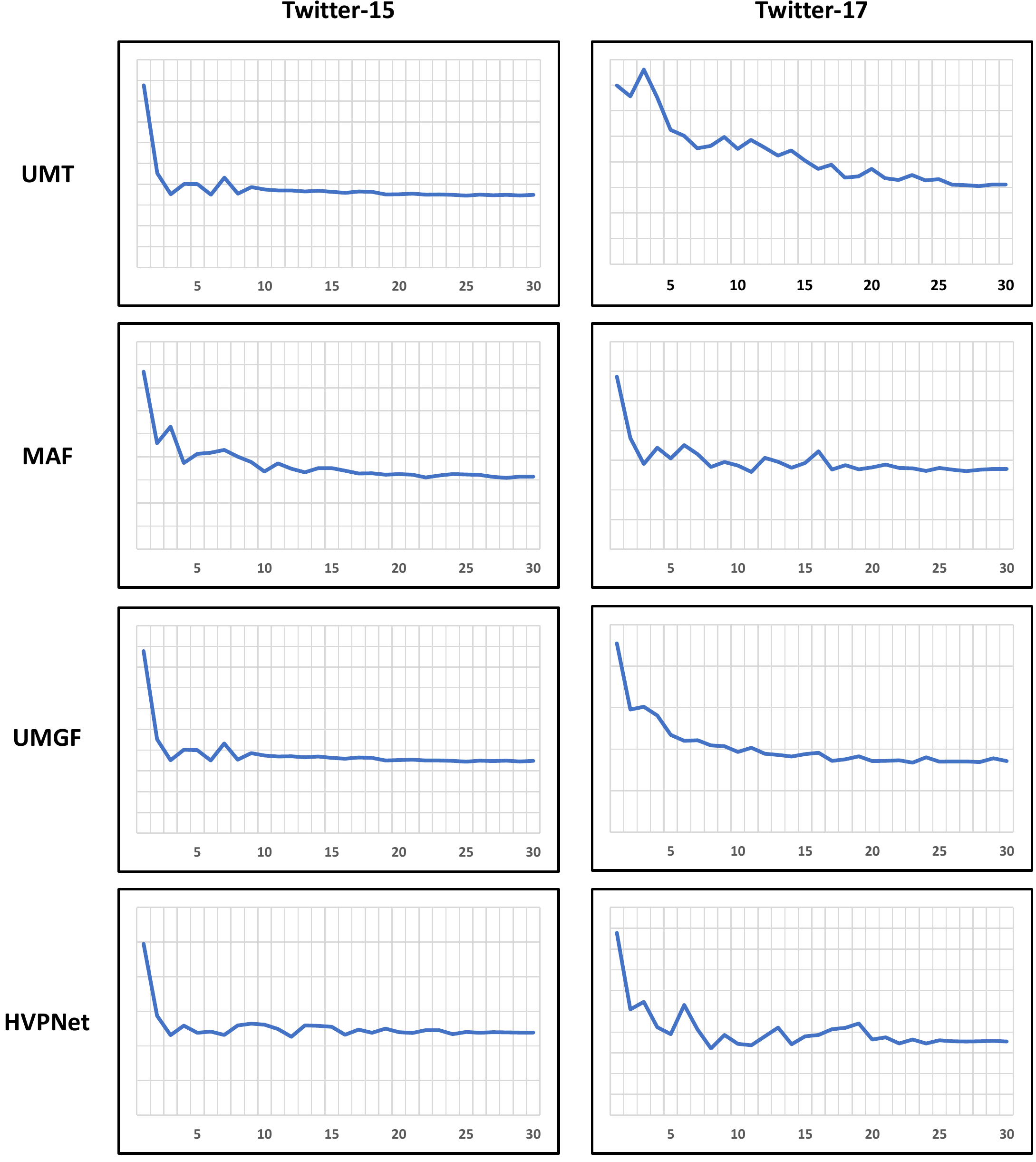}
    \caption{The uncertainty estimation. We gradually increased the training epochs (X-axis) and observed a gradual decrease in epistemic uncertainty (Y-axis) in the test dataset, which eventually reached a stable level. Lower levels of epistemic uncertainty indicate a more stable and confident model.}
    \label{Figure: uncertainty}
    \vspace{-0.3cm}
\end{figure}

%% file: 5_Conclusion.tex
\section{Conclusion}
In this work, we propose a novel algorithm for multimodal named entity recognition (MNER) by incorporating uncertainty estimation. The algorithm models uncertainty under a fully probabilistic framework and utilizes normal-inverse gamma distributions (NIG) to fuse multiple modality information. The algorithm aims to promote both prediction accuracy and trustworthiness. By exploring the potential of pre-trained large foundation models in MNER, we further improve the performance.
Experiments on two datasets and comparisons with four baseline models validate the effectiveness, robustness, and reliability of the proposed model, outperforming the state-of-the-art and achieving new state-of-the-art results. 